\documentclass[11pt]{article}

\usepackage[preprint]{acl}

\usepackage{times}
\usepackage{latexsym}
\usepackage{amsmath}
\usepackage{algorithm}
\usepackage{algpseudocode}
\usepackage{booktabs}
\usepackage{multirow}
\usepackage{colortbl}
\definecolor{oursrow}{gray}{0.93}
\usepackage[T1]{fontenc}

\usepackage[utf8]{inputenc}
\usepackage{newunicodechar}
\newunicodechar{【}{[}
\newunicodechar{】}{]}

\usepackage{microtype}

\usepackage{inconsolata}

\usepackage{graphicx}

\title{Branch and Consolidate: A Unified Recipe for Multi-Teacher On-Policy Distillation}

\author{%
  \textbf{Zihan Liang}\textsuperscript{1,5}%
  \thanks{These authors contributed equally.}\qquad
  \textbf{Zikun Qu}\textsuperscript{1,2}\footnotemark[1]%
  \thanks{Work done during an internship at Kuaishou Technology.}\qquad
  \textbf{Ben Chen}\textsuperscript{1}\footnotemark[1]%
  \thanks{Co-corresponding authors.}\qquad
  \textbf{Fan Yang}\textsuperscript{1}\\
  \textbf{Wenwu Ou}\textsuperscript{1}\qquad
  \textbf{Min Zhang}\textsuperscript{4}\footnotemark[3]\qquad
  \textbf{Zhongxiang Dai}\textsuperscript{2}\footnotemark[3]\qquad
  \textbf{Xinchao Wang}\textsuperscript{3}\\
  \textsuperscript{1}Kuaishou Technology\qquad
  \textsuperscript{2}CUHKSZ\qquad
  \textsuperscript{3}NUS\qquad
  \textsuperscript{4}ECNU\qquad
  \textsuperscript{5}UPenn\\
  \texttt{benchen4395@gmail.com}%
}

\begin{document}
\maketitle
\begin{abstract}
Specialists post-trained from a shared checkpoint acquire complementary capabilities, and multi-teacher on-policy distillation (MOPD) can consolidate them into a single model.
However, we find that existing consolidation methods, from weight merging to MOPD, fall short when specialists diverge substantially from the shared checkpoint, whereas prior studies mostly examine specialists that remain close to it.
We present a unified recipe for multimodal MOPD that consolidates such specialists back into the single student they branched from.
The recipe specifies initialization, data selection, balanced supervision across teachers, and preservation of existing capabilities for consolidation, together with a targeted refinement stage for capabilities that still lag behind their specialists.
In an industrial setting with six specialists, the student closely approaches each specialist in its own domain and surpasses the perception specialist. It recovers 97.0\% of their average gain, compared with 79.8\% for IM-MOPD.
In the fully open five-domain setting with Qwen3-4B, our method recovers 95.7\% of the specialists' average gain after consolidation and 97.5\% after refinement, compared with 94.1\% for IM-MOPD.
We further analyze the full pipeline through mechanism ablations and training dynamics, providing practical guidance for diagnosing incomplete consolidation. 
Code will be released.
\end{abstract}

\section{Introduction}
\par
Frontier post-training increasingly trains specialists independently from a shared checkpoint and then consolidates them into a single model~\citep{deepseekai2026deepseekv4highlyefficientmilliontoken,kimiteam2026kimik3openfrontier,nvidia2026nemotron3ultraopen}.
A common tool for this consolidation is multi-teacher on-policy distillation (MOPD), in which the student generates its own responses and each specialist supervises those in its own domain~\citep{coreteam2026mimov2flashtechnicalreport,ma2026mopdmultiteacheronpolicydistillation}. 
Industrial reports describe the pipelines they adopt but offer little systematic analysis of why design choices work or how failures should be diagnosed.

\begin{figure*}[!t]
  \centering
  \includegraphics[width=\textwidth]{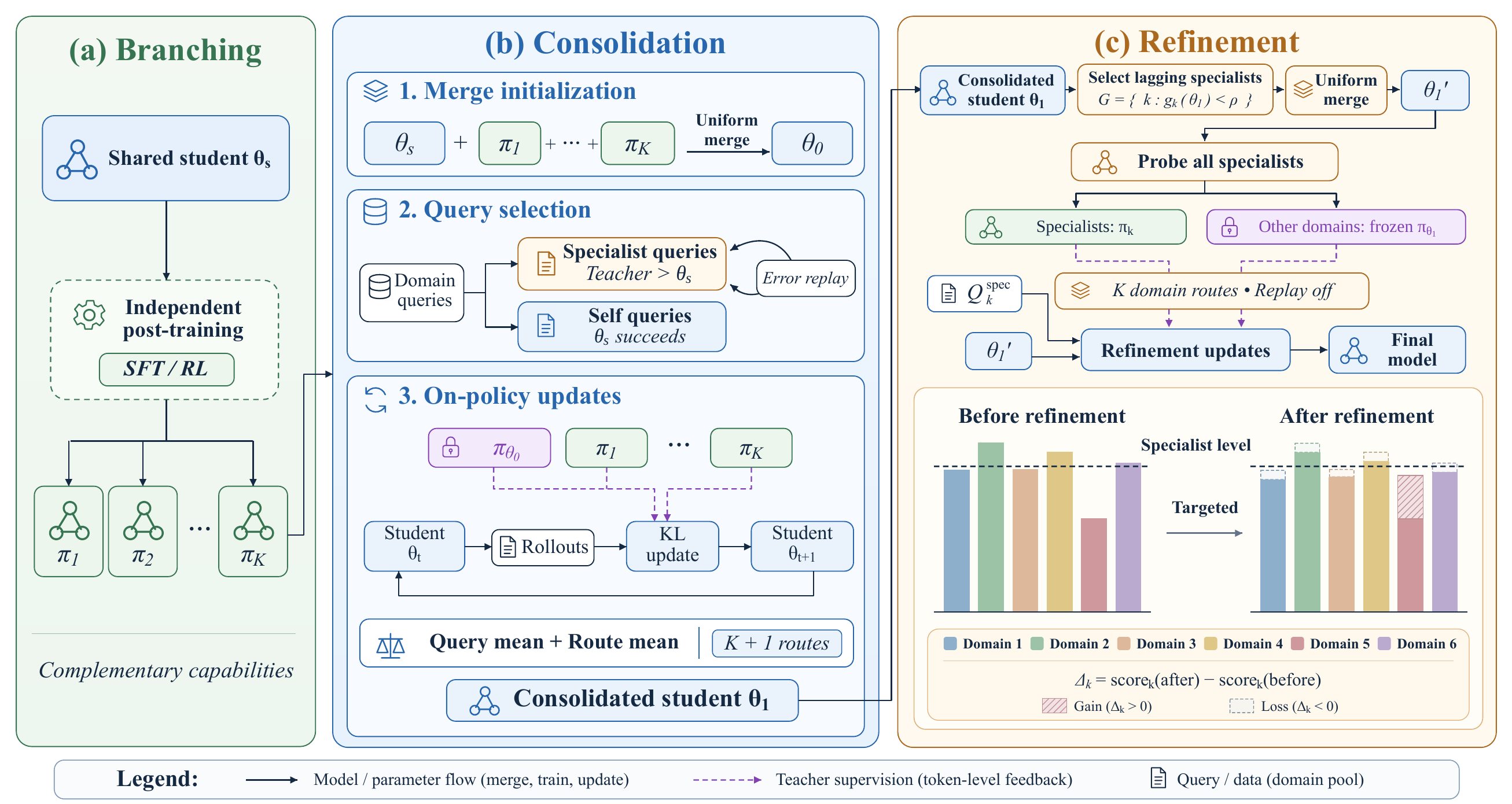}
  \caption{Overview of our recipe. (a) Specialists are post-trained independently from a shared student. (b) They are consolidated back into the student through MOPD. (c) The student is then refined where capabilities still lag behind.}
  \label{fig:overview}
\end{figure*}

\par
Public studies dedicated to consolidation provide such analysis through controlled experiments~\citep{gao2026openmopddiagnosingfixingcapability,sun2026d3mopddynamicdomainscheduling,yin2026slowfastmultiteacheronpolicydistillation,li2026teacherassignmentdomainnormalizedmultiteacher,kim2026paingainiterativemerging}.
Most of them, however, obtain specialists by single-domain reinforcement learning from a shared checkpoint, which tends to keep the specialists close to the student. 
In this regime consolidation is comparatively easy, and even training-free weight merging or well-tuned offline fine-tuning recovers most specialist gains~\citep{kaiyuan2026fishergeometricviewmultiteacheronpolicy,xie2026rethinkingselfdistillationmultiteachercapability}.
Practical specialists are often produced by multi-stage pipelines that combine continued pretraining~\citep{grattafiori2024llama3herdmodels,sukhbaatar2024branchtrainmixmixingexpertllms,deepseekai2024deepseekcoderv2breakingbarrierclosedsource}, supervised fine-tuning, and reinforcement learning~\citep{deepseekai2026deepseekv4highlyefficientmilliontoken,coreteam2026mimov2flashtechnicalreport,5team2025glm45agenticreasoningcoding}, and can therefore drift far from the student. 
We find that existing methods fall short in this regime, especially when the overlap ratio between the top-ranked tokens of teacher and student is low~\citep{li2026rethinkingonpolicydistillationlarge}.

\par
We trace this shortfall to four sources, namely an initialization from which teacher supervision is hard to exploit, training queries that reduce the distillation loss faster than they add capability, a loss aggregation that silently reallocates supervision across specialists, and the lack of an anchor that preserves what the student already knows.
We therefore present a unified recipe that addresses them jointly (Figure~\ref{fig:overview}). 
The student is initialized by merging it with the specialists~\citep{kim2026paingainiterativemerging}, which falls short as a final model but raises the overlap, and is distilled on queries that a specialist solves but the student does not.
Its loss is averaged first over the queries of each specialist and then equally across specialists~\citep{bai2026scalinghorizonparametersreaching}, so that neither response length nor dataset size determines a specialist's weight in the objective. 
A frozen copy of the merged initialization serves as an additional teacher that anchors the student near its starting point during consolidation, much like a KL penalty.
After consolidation, a targeted refinement stage closes the remaining gaps while minimizing the erosion of capabilities already recovered.

\par
In an industrial setting with six substantially diverged specialists, the resulting student closely approaches every specialist in its own domain and surpasses one of them, recovering 97.0\% of their average gain over the student, compared with 79.8\% for the strongest prior method~\citep{kim2026paingainiterativemerging} and 42.3\% for the best weight-merging method.
In the fully open five-domain setting of IM-MOPD with Qwen3-4B, whose specialists also start from a low overlap with the student, the recipe recovers 97.5\% of the specialists' average gain, compared with 94.1\% for IM-MOPD (Appendix~\ref{app:open}).
Beyond the main results, we re-examine prescriptions from prior studies in Section~\ref{sec:design}, holding the specialists and training budget fixed, and find that several fall short once specialists diverge.
We further diagnose how consolidation unfolds in Section~\ref{sec:diagnosis}, and collect practical observations on reading training signals, allocating the training budget, and configuring refinement in Appendix~\ref{app:monitor}.

\par
Our contributions are as follows.
\begin{itemize}
  \item We show that existing consolidation methods fall short when specialists diverge substantially from their shared checkpoint, a regime that public studies rarely cover.
  \item We present a unified recipe for multimodal MOPD that consolidates specialists back into the student they branched from and outperforms prior methods in both industrial and fully open settings.
  \item We re-examine design choices across initialization, data, loss aggregation, objectives, preservation, and refinement under controlled conditions, and provide a practical guide for diagnosing incomplete consolidation.
\end{itemize}


\section{Related Work}
\label{sec:related}

\paragraph{Multi-teacher on-policy distillation.}
On-policy distillation trains a student on its own samples under token-level teacher supervision~\citep{agarwal2024onpolicydistillationlanguagemodels,gu2026minillmonpolicydistillationlarge}, which reduces the train--inference mismatch of offline distillation~\citep{hinton2015distillingknowledgeneuralnetwork}.
Frontier reports adopt its multi-teacher form to consolidate domain specialists~\citep{deepseekai2026deepseekv4highlyefficientmilliontoken,coreteam2026mimov2flashtechnicalreport,nvidia2026nemotron3ultraopen,kimiteam2026kimik3openfrontier}.
Public studies propose individual remedies, including rebalancing supervision across domains~\citep{gao2026openmopddiagnosingfixingcapability,li2026teacherassignmentdomainnormalizedmultiteacher}, scheduling domains dynamically~\citep{sun2026d3mopddynamicdomainscheduling}, separating fast and slow updates~\citep{yin2026slowfastmultiteacheronpolicydistillation}, merging specialists during training~\citep{kim2026paingainiterativemerging}, aggregating the loss per teacher~\citep{bai2026scalinghorizonparametersreaching}, and modeling the tail of the teacher distribution~\citep{huang2026tailawaretopkonpolicydistillation}.
These studies mostly use specialists that stay close to the shared checkpoint and examine each remedy in isolation.
We instead study substantially diverged specialists and re-examine these remedies jointly under a fixed budget.

\paragraph{Model merging.}
Weight-space methods combine fine-tuned models without further training, by averaging~\citep{wortsman2022modelsoupsaveragingweights}, adding task vectors~\citep{ilharco2023editingmodelstaskarithmetic}, or resolving parameter interference~\citep{yadav2023tiesmergingresolvinginterferencemerging}.
Branch-train-merge pipelines similarly train experts in parallel before combining them~\citep{li2022branchtrainmergeembarrassinglyparalleltraining,sukhbaatar2024branchtrainmixmixingexpertllms}.
Merging works well when models stay within a shared basin~\citep{kaiyuan2026fishergeometricviewmultiteacheronpolicy}, but its gains on large language models are less reliable~\citep{hitit2026systematicstudyinthewildmodel}.
We use merging only to initialize distillation, where it raises the teacher--student overlap.

\paragraph{Offline distillation and forgetting.}
Offline alternatives fine-tune the student on specialist responses filtered for correctness, which can recover most of the gain when specialists stay close to the student~\citep{xie2026rethinkingselfdistillationmultiteachercapability}.
Distillation toward new teachers can also erode what the student already knows, which Nemotron~3 Ultra counters by using an earlier checkpoint of the student as a self teacher for areas that no specialist covers~\citep{nvidia2026nemotron3ultraopen}.
Our frozen self teacher follows this idea and anchors the student to its merged initialization.

\section{Method}
\label{sec:method}

\par
This section presents our recipe, summarized in Figure~\ref{fig:overview}.
We introduce the notation and the MOPD objective in Section~\ref{sec:setup} and describe how specialists are consolidated back into the student in Section~\ref{sec:consolidation}.
Section~\ref{sec:refinement} explains how refinement targets the remaining gaps. Algorithm~\ref{alg:recipe} in Appendix~\ref{app:method} gives the full procedure.

\subsection{Preliminaries}
\label{sec:setup}

\par
Let $\theta_s$ denote the student checkpoint.
From it, $K$ specialists $\{\pi_k\}_{k=1}^{K}$ with parameters $\{\phi_k\}_{k=1}^{K}$ are branched, each post-trained for one capability through its own pipeline, which may combine large-scale supervised fine-tuning and reinforcement learning.
Specialist $k$ is associated with a query distribution $\mathcal{D}_k$ for its capability, where a query may contain text, images, or videos.
All specialists share the architecture and tokenizer of the student, and the goal is a single model of the same size that retains the capability of every specialist.

\par
Each query $x \sim \mathcal{D}_k$ is routed to its specialist $k(x)$.
The student samples a response from a snapshot $\pi_{\bar\theta}$ of its parameters taken before each update.
Writing $p_{x,t} = \pi_\theta(\cdot \mid h_{x,t})$ and $q_{x,t} = \pi_{k(x)}(\cdot \mid h_{x,t})$ for the student and teacher distributions at the prefix $h_{x,t}$ of each generated position, the student minimizes
\begin{equation}
\ell_{x,t}(\theta) = D_{\mathrm{KL}}\bigl(p_{x,t} \,\big\|\, q_{x,t}\bigr).
\label{eq:mopd}
\end{equation}
For agentic queries, $h_{x,t}$ includes earlier tool observations, which are excluded from the loss.
We backpropagate this divergence directly instead of using the per-token log-ratio as a policy-gradient advantage, and update once per rollout batch.

\par
We characterize teacher--student compatibility on student-generated prefixes using the top-$b$ token overlap ratio,
\begin{equation}
o_{x,t} =
\frac{
\bigl|\operatorname{Top}_b(p_{x,t})
\cap
\operatorname{Top}_b(q_{x,t})\bigr|
}{b},
\label{eq:overlap}
\end{equation}
where $\operatorname{Top}_b(r)$ denotes the set of the $b$ most probable tokens under distribution $r$.
We set $b=32$ and average the overlap over generated tokens.
Low overlap indicates that the teacher and student favor different next-token candidates at the same prefix, where the teacher's supervision is harder to exploit.
We use this metric to compare student initializations, select teachers for refinement, and analyze consolidation dynamics.

\subsection{Consolidation}
\label{sec:consolidation}

\par
The consolidation stage combines four components, a merged initialization, query selection with error replay, a self teacher, and a balanced objective, illustrated in Figure~\ref{fig:overview}b and analyzed individually in Section~\ref{sec:design}.

\paragraph{Merge initialization.}
Because the specialists have drifted from the student checkpoint, distillation from it starts at a low overlap.
We therefore initialize the student by uniformly averaging all parameters of the student checkpoint and all $K$ specialists,
\begin{equation}
\theta_0 = \frac{1}{K+1}\Bigl(\theta_s + \sum_{k=1}^{K}\phi_k\Bigr).
\label{eq:merge}
\end{equation}
Merging alone recovers only part of the specialists' gains, but raises teacher--student overlap and provides a more effective starting point for subsequent MOPD training.

\paragraph{Query selection.}
We use the student checkpoint $\theta_s$ as the fixed reference for query selection.
We construct each specialist query pool $\mathcal Q_k^{\mathrm{spec}}$ by selecting queries on which $\pi_k$ outperforms the reference student.
For tasks with verifiable answers, the specialist and the reference student each answer once, and we retain queries on which only the specialist is correct.
For open-ended tasks scored by rubrics, such as captioning, we keep the queries with the largest score gap between the two.
We construct a separate self-query pool \(\mathcal Q_k^{\mathrm{self}}\) for each domain by randomly sampling from its unfiltered candidate pool and retaining queries on which the reference student succeeds.
Their union, \(\mathcal Q^{\mathrm{self}}=\bigcup_{k=1}^{K}\mathcal Q_k^{\mathrm{self}}\), supplies the self-teacher route.
From the start of training, queries that the current student fails enter an error pool for their route.
After the first quarter of training, each specialist route draws a fixed number of new queries per batch plus a varying number of queries replayed from its pool, which we call error replay.
Replayed queries receive fresh rollouts and leave the pool once they are answered correctly.

\paragraph{Self teacher.}
A frozen copy of the merged initialization, $\pi_0 = \pi_{\theta_0}$, serves as an additional teacher with its own route, anchoring the student near its starting point much like a KL penalty.
It supervises the self queries described above and receives the same weight as each specialist in the objective below.
Its role is to limit drift from the initialization rather than to supply correct answers, so the self queries are chosen by the student checkpoint $\theta_s$.

\paragraph{Objective and aggregation.}
Evaluating Eq.~\eqref{eq:mopd} over the full vocabulary is costly, so we approximate it on the $b$ tokens ranked highest by the teacher, $S = \mathrm{Top}_b(q_{x,t})$, plus a tail bucket that collects the remaining probability mass~\citep{huang2026tailawaretopkonpolicydistillation}.
Dropping the subscripts, the per-position loss is
\begin{equation}
\hat{\ell}_{x,t} = \sum_{v \in S} p(v) \log \frac{p(v)}{q(v)} + \bar{p} \log \frac{\bar{p}}{\bar{q}},
\label{eq:topk}
\end{equation}
where $\bar{p}$ and $\bar{q}$ are the probabilities that the student and the teacher assign outside $S$.

\par
Routes differ widely in response length, and error replay makes the number of queries each batch draws from a route vary over training.
Averaging $\hat{\ell}$ over all tokens would weight each route by both, so we average over the queries within each route and then over routes,
\begin{equation}
\mathcal{L} = \frac{1}{|\mathcal{R}|} \sum_{k \in \mathcal{R}} \frac{1}{|\mathcal{B}_k|} \sum_{x \in \mathcal{B}_k} \frac{1}{|T_x|} \sum_{t \in T_x} \hat{\ell}_{x,t},
\label{eq:agg}
\end{equation}
where $\mathcal{R}$ is the set of teacher routes, $\mathcal{B}_k$ is the set of queries in the batch routed to teacher $k$, and $T_x$ is the set of generated positions of query $x$.
When every route contributes the same number of queries, the outer average reduces to a plain mean over queries, so it matters only when routes are sampled unevenly.
In consolidation, $\mathcal{R}$ contains the $K$ specialists and the self teacher.

\subsection{Refinement}
\label{sec:refinement}
Consolidation can still leave some capabilities far behind their specialists. 
Let $\theta_1$ denote the student after consolidation, and let $a_k(\theta)$ denote the held-out validation score of $\pi_\theta$ for capability $k$.
We measure recovery relative to the original student checkpoint
$\theta_s$:
\begin{equation}
g_k(\theta_1)=
\frac{a_k(\theta_1)-a_k(\theta_s)}
{a_k(\phi_k)-a_k(\theta_s)}.
\label{eq:recovery}
\end{equation}
The set of specialists corresponding to lagging capabilities is $\mathcal G=\{k:g_k(\theta_1)<\rho\}$.
We initialize refinement by uniformly averaging the parameters of the consolidated student and all specialists in $\mathcal G$:
\begin{equation}
\theta_1'=
\frac{1}{1+|\mathcal G|}
\Bigl(\theta_1+\sum_{k\in\mathcal G}\phi_k\Bigr).
\label{eq:remerge}
\end{equation}
This gives $\theta_1'=\theta_1$ when $\mathcal G$ is empty.
Because merging the lagging specialists can lower the overlap with the others, we then reassess compatibility with all specialists using the overlap ratio in Eq.~\eqref{eq:overlap}.
Let $\bar o_k(\theta_1')$ denote the mean overlap between the merged student $\pi_{\theta_1'}$ and specialist $\pi_k$, averaged over generated tokens on a fixed probe set for route $k$.
We select
\[
\mathcal T=\mathcal G\cup
\{k\notin\mathcal G:\bar o_k(\theta_1')<\tau\}
\]
as the specialist teachers for subsequent updates.
Each domain retains its own route, taught by $\pi_k$ if $k\in\mathcal T$ and otherwise by a frozen copy of $\pi_{\theta_1}$ on its self pool $\mathcal Q_k^{\mathrm{self}}$.
Starting from $\theta_1'$, we perform a short MOPD run with this teacher assignment.
We use the objective in Eq.~\eqref{eq:agg} with $\mathcal R=\{1,\ldots,K\}$.
Specialist routes reuse their consolidation query pools, and error replay is disabled.

\section{Experiments}
\label{sec:experiments}

\par
We evaluate the recipe in an industrial setting and in a fully open setting, then re-examine its design choices and diagnose consolidation dynamics.

\subsection{Experimental Setup}
\label{sec:exp_setup}

\paragraph{Student and specialists.}
The student is an internal checkpoint obtained by continued pretraining of Qwen3.6-35B-A3B~\citep{qwen36_35b_a3b}.
All six specialists start from this checkpoint and are post-trained with large-scale supervised fine-tuning followed by reinforcement learning, one for each of instruction following, perception, agentic business tasks, business understanding, general reasoning, and domain reasoning.
Queries may contain images or videos, and agentic business tasks require multi-turn tool use.
The candidate query pool of each capability reuses the queries from the supervised fine-tuning or reinforcement learning stages of its specialist, and both $\mathcal{Q}_k^{\mathrm{spec}}$ and $\mathcal{Q}_k^{\mathrm{self}}$ are drawn from it.
Table~\ref{tab:routes} reports the average response length of each route and its initial overlap when distillation starts from the student or from the merged initialization.

\begin{table}[t]
\centering
\small
\setlength{\tabcolsep}{4pt}
\begin{tabular}{lrccc}
\toprule
\multirow{2}{*}{Route} & \multirow{2}{*}{Avg. tokens} & \multicolumn{3}{c}{Initial overlap (\%)} \\
\cmidrule(lr){3-5}
 & & $\theta_s$ & $\theta_0$ & $\Delta$ \\
\midrule
Instruction following & 2{,}800 & 77.6 & 83.2 & +5.6 \\
Perception            & 1{,}100 & 85.5 & 88.0 & +2.5 \\
Agentic business      & 5{,}400 & 75.8 & 80.7 & +4.9 \\
Business underst.     & 2{,}000 & 78.7 & 83.6 & +4.9 \\
General reasoning     & 5{,}500 & 68.8 & 75.3 & +6.5 \\
Domain reasoning      & 1{,}300 & 79.8 & 84.9 & +5.1 \\
\bottomrule
\end{tabular}
\caption{Per-route average response length of the student $\theta_s$ and initial overlap with the specialist when distillation starts from $\theta_s$ or from the merged initialization $\theta_0$, and the gain $\Delta$ from merging.}
\label{tab:routes}
\end{table}

\paragraph{Evaluation.}
Each capability is evaluated on a group of benchmarks.
Instruction following uses IFEval~\citep{zhou2023instructionfollowingevaluationlargelanguage}, MM-IFEval~\citep{ding2025mmifenginemultimodalinstructionfollowing}, and MIA-Bench~\citep{qian2025miabenchbetterinstructionfollowing} together with internal benchmarks, and general reasoning uses IMO-AnswerBench~\citep{luong-etal-2025-towards}, MathVision~\citep{wang2024measuringmultimodalmathematicalreasoning}, LogicVista~\citep{xiao2024logicvistamultimodalllmlogical}, and the ten-option and vision settings of MMMU-Pro~\citep{yue2025mmmuprorobustmultidisciplinemultimodal}.
Perception, agentic business tasks, business understanding, and domain reasoning are evaluated on internal benchmarks, which we do not disclose for proprietary reasons.
The score of a capability averages its benchmarks.

\paragraph{Metrics.}
Let $S_k(\theta)$ denote the score of $\pi_\theta$ on capability $k$ and $\bar{S}(\theta)$ its mean over the $K$ capabilities.
Our primary metric is the recovery rate
\begin{equation}
\mathrm{Rec}(\theta) = \frac{\bar{S}(\theta) - \bar{S}(\theta_s)}{\bar{S}_{\mathrm{spec}} - \bar{S}(\theta_s)},
\label{eq:rec}
\end{equation}
where $\bar{S}_{\mathrm{spec}}$ averages $S_k(\phi_k)$ over capabilities, so that 100\% means matching the specialists on average.
We compute it from aggregate gains rather than averaging per-capability ratios, since some specialists improve only slightly on the student and per-capability ratios would amplify small fluctuations.
To show where a model falls short, we also report the normalized score $S_k(\theta)/S_k(\phi_k)$ of each capability, so that 100\% corresponds to the specialist.
Both metrics are computed on the evaluation benchmarks, unlike $g_k$ in Section~\ref{sec:refinement}, which is computed per capability on the validation set and serves only as a coarse threshold to flag capabilities that lag far behind.

\paragraph{Baselines.}
We compare against three families of consolidation methods.
Weight merging includes uniform averaging, which equals our initialization alone, Task Arithmetic~\citep{ilharco2023editingmodelstaskarithmetic}, and TIES~\citep{yadav2023tiesmergingresolvinginterferencemerging}.
Offline distillation includes RS-SFT, which fine-tunes the student on specialist responses judged correct~\citep{xie2026rethinkingselfdistillationmultiteachercapability}, and offline KD, which applies our objective to specialist responses.
On-policy methods include vanilla MOPD~\citep{ma2026mopdmultiteacheronpolicydistillation}, which uses our objective but samples queries at random and averages the loss over all tokens without a self teacher, and four variants, Open-MOPD~\citep{gao2026openmopddiagnosingfixingcapability}, D$^3$-MOPD~\citep{sun2026d3mopddynamicdomainscheduling}, SF-MOPD~\citep{yin2026slowfastmultiteacheronpolicydistillation}, and IM-MOPD~\citep{kim2026paingainiterativemerging}, each implemented by adding its core mechanism to vanilla MOPD.
All methods use the same candidate query pools and training budget, with checkpoints selected on the validation set (Appendix~\ref{app:baselines}).

\begin{table*}[t]
\centering
\small
\begin{tabular}{lcccccccc}
\toprule
\multirow{2}{*}{Method} & \multicolumn{6}{c}{Capability} & \multicolumn{2}{c}{Overall} \\
\cmidrule(lr){2-7} \cmidrule(lr){8-9}
 & IF & Perc. & Agent & Bus. & Gen. & Dom. & Avg. & Rec. (\%) \\
\midrule
Qwen3.6-35B-A3B & 70.25 & 56.41 & 54.01 & 71.68 & 77.07 & 39.38 & 61.47 & -- \\
Student & 71.70 & 59.30 & 72.90 & 76.94 & 75.26 & 64.50 & 70.10 & 0.0 \\
Specialists & 75.73 & 61.63 & 74.56 & 81.33 & 79.22 & 69.57 & 73.67 & 100.0 \\
\midrule
\multicolumn{9}{l}{\textit{Weight merging}} \\
Weight averaging & 73.04 & 60.11 & 71.94 & 80.21 & 74.40 & 66.07 & 70.96 & 24.1 \\
Task Arithmetic & 73.51 & 60.58 & 72.40 & 80.67 & 74.86 & 66.53 & 71.42 & 37.1 \\
TIES & 72.54 & 60.22 & 72.20 & \textbf{83.06} & 75.15 & 66.50 & 71.61 & 42.3 \\
\midrule
\multicolumn{9}{l}{\textit{Offline distillation}} \\
RS-SFT & 72.61 & 59.83 & 70.70 & 79.98 & 74.12 & 65.65 & 70.48 & 10.7  \\
Offline KD & 73.31 & 59.88 & 73.23 & 79.35 & 73.48 & 66.53 & 70.96 & 24.2 \\
\midrule
\multicolumn{9}{l}{\textit{On-policy distillation}} \\
Vanilla MOPD & 73.37 & 60.68 & 69.26 & 81.51 & 75.87 & 66.34 & 71.17 & 30.0 \\
Open-MOPD & 74.77 & 61.23 & 73.43 & 81.46 & 76.29 & 68.26 & 72.57 & 69.2 \\
D$^3$-MOPD & 74.58 & 61.11 & 73.54 & 81.02 & 76.46 & 68.01 & 72.45 & 65.9 \\
SF-MOPD & 73.44 & 60.46 & 72.56 & 80.19 & 74.92 & 66.58 & 71.36 & 35.2 \\
IM-MOPD & 75.09 & 61.41 & 73.49 & 81.49 & \underline{77.57} & 68.66 & 72.95 & 79.8 \\
\midrule
\multicolumn{9}{l}{\textit{Ours}} \\
\rowcolor{oursrow} Consolidation & \underline{75.47} & \underline{62.20} & \textbf{74.38} & \underline{81.68} & 76.76 & \textbf{69.69} & \underline{73.36} & \underline{91.3} \\
\rowcolor{oursrow} Full recipe & \textbf{75.60} & \textbf{62.90} & \underline{74.30} & 80.90 & \textbf{78.90} & \underline{68.80} & \textbf{73.57} & \textbf{97.0} \\
\bottomrule
\end{tabular}
\caption{Main results. Rec. is the recovery rate in Eq.~\eqref{eq:rec}, the share of the specialists' average gain over the student that each model recovers. Best results are in bold and second best underlined.}
\label{tab:main}
\end{table*}

\paragraph{Implementation.}
Responses are judged by Qwen3.8-27B~\citep{qwen38} throughout.
Consolidation runs for 164 updates with a batch of 256 new queries, about one sixth of them from $\mathcal{Q}^{\mathrm{self}}$, and a constant learning rate of $10^{-6}$.
After the first quarter of training, error replay adds up to as many replayed queries as new ones on each specialist route.
The consolidation checkpoint $\theta_1$ is selected on the validation set.
Refinement starts from $\theta_1'$ with recovery threshold $\rho = 0.5$ and overlap threshold $\tau = 0.8$ and runs for 10 updates with the same batch size and learning rate, half of each batch from $\mathcal{Q}^{\mathrm{self}}$.
Other settings are given in Appendix~\ref{app:setup} and training costs in Appendix~\ref{app:cost}.

\begin{figure*}[t]
\centering
\includegraphics[width=\textwidth]{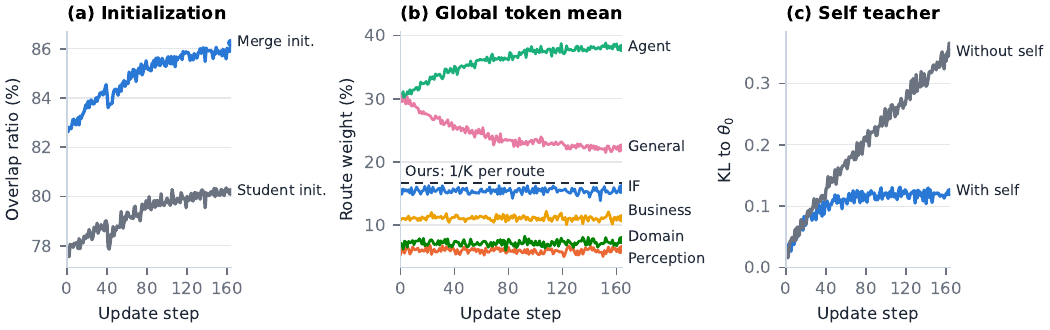}
\caption{Training dynamics behind three design choices. (a) Merging raises the overlap from the start. (b) Under a global token mean, the shares of the specialist routes follow response lengths and drift during training, while our aggregation keeps each at $1/K$. (c) The self teacher bounds the divergence from $\theta_0$.}
\label{fig:design}
\end{figure*}

\subsection{Main Results}
\label{sec:main_results}
Table~\ref{tab:main} reports the results, with per-benchmark general reasoning scores in Appendix~\ref{app:per_bench}.
The full recipe recovers 97.0\% of the specialists' average gain, compared with 79.8\% for IM-MOPD, the strongest prior method, and 42.3\% for TIES, the best weight-merging method.
Its normalized score is at least 98.9\% on every capability, and it surpasses the perception specialist.
Consolidation alone recovers 91.3\%, above all baselines.

\par
Vanilla MOPD recovers only 30.0\%, below TIES, and even falls below the student on agentic business tasks.
Offline KD, which applies the same objective to specialist responses, recovers 24.2\% and RS-SFT only 10.7\%, so on-policy sampling helps but is not sufficient on its own.
Among the variants without merging, rebalancing or rescheduling domains brings the largest gains, with Open-MOPD and D$^3$-MOPD reaching 69.2\% and 65.9\%, whereas SF-MOPD reaches 35.2\%.
IM-MOPD, which starts from a merged initialization that raises the initial overlap (Table~\ref{tab:routes}), reaches 79.8\%.

\par
Refinement targets general reasoning, the only capability with $g_k < \rho$ after consolidation.
In 10 updates it raises general reasoning from 96.9\% to 99.6\% of the specialist, while business understanding and domain reasoning drop by about one point and stay close to their specialists.
Refinement costs 35 GPU hours, compared with 550 for consolidation (Appendix~\ref{app:cost}).

\par
We also evaluate the recipe in the fully open setting of IM-MOPD (Appendix~\ref{app:open}), where five specialists are branched from Qwen3-4B-Base fine-tuned on OpenThoughts3.
We choose this setting deliberately, since three of its five specialists start below 70\% overlap with the student, the regime our recipe targets (Table~\ref{tab:open_overlap}).
Our recipe recovers 95.7\% of the specialists' average gain after consolidation and 97.5\% after refinement, compared with 94.1\% for IM-MOPD. The refined student outperforms IM-MOPD on all five benchmarks (Table~\ref{tab:open_results}).

\begin{figure*}[t]
\centering
\includegraphics[width=\textwidth]{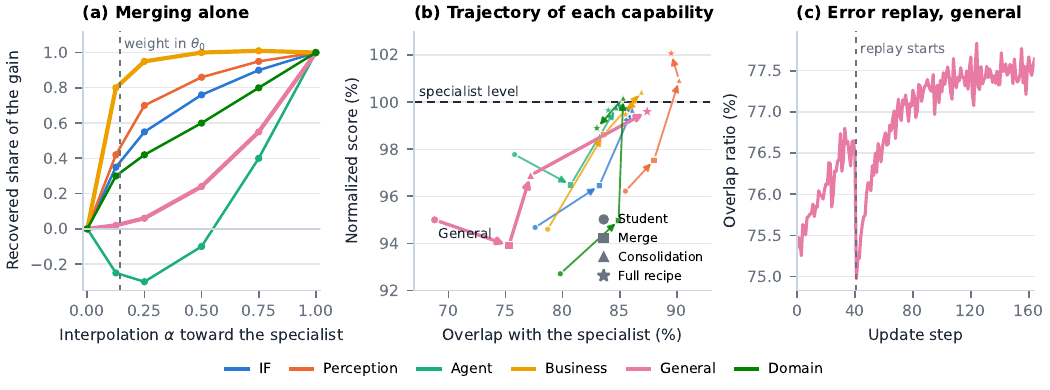}
\caption{Diagnosis of consolidation. (a) Recovered share of each specialist's gain when interpolating from the student toward the specialist; the dashed line marks the weight of each specialist in $\theta_0$. (b) Overlap and normalized score of each capability from the student through merging and consolidation to the full recipe. (c) Overlap on general reasoning drops when error replay starts and then recovers.}
\label{fig:diag}
\end{figure*}

\subsection{Design Choices}
\label{sec:design}
We ablate each consolidation component under a fixed training budget (Table~\ref{tab:ablation}) and show the dynamics of three components in Figure~\ref{fig:design}.

\paragraph{Initialization.}
Starting from the student instead of the merged initialization lowers recovery from 91.3\% to 71.4\%, and initializing with RS-SFT before consolidation recovers 75.1\%.
Even from the student, the remaining components outperform every MOPD variant without merging in Table~\ref{tab:main}, and merging adds about 20 points on top of them.
As shown in Figure~\ref{fig:design}a, merging raises the overlap from the first update, so distillation starts closer to every specialist.

\paragraph{Training queries.}
Replacing selected queries with random ones lowers teacher loss and raises overlap, but costs 13.9 recovery points (Appendix~\ref{app:design}).
The student and specialists likely agree on most random queries, leaving little training signal.
Even a tenfold larger random pool trails by 7.8 points. Removing error replay costs 6.0 points.

\paragraph{Loss aggregation.}
A global token mean loses 30.6 points, since route weights then follow response lengths and drift during training rather than reflecting which capability needs training (Figure~\ref{fig:design}b).
Agentic business tasks receive the largest and a growing share yet fall below the student, as in vanilla MOPD, which uses the same aggregation (Table~\ref{tab:main}), so more weight on long trajectories does not translate into more capability.
Averaging tokens within each route still loses 15.3 points.
Mechanisms from prior work lose 2.1 to 12.5 points: token-share sampling~\citep{gao2026openmopddiagnosingfixingcapability} gives routes with long responses, such as general reasoning and agentic tasks, fewer queries, dynamic sampling~\citep{sun2026d3mopddynamicdomainscheduling} shifts budget toward general reasoning at the expense of the other routes, and spread normalization loses the most.

\paragraph{Objective and preservation.}
Dropping the tail bucket, restricting the loss to overlapping candidates, or using only the sampled token each loses 15.8 to 18.7 points, and with the sampled token only, the entropy of the student rises from about 0.5 to 0.65.
Without the self teacher, recovery drops by 30.5 points.
With it, the KL to $\theta_0$ levels off at about 0.12, whereas without it the KL keeps rising beyond 0.3 (Figure~\ref{fig:design}c), so the self teacher bounds drift without blocking learning.
Further ablations and the per-capability scores of every variant are given in Appendix~\ref{app:design}.

\begin{table}[t]
\centering
\footnotesize
\setlength{\tabcolsep}{4pt}
\begin{tabular}{lrr}
\toprule
Variant & Rec. (\%) & $\Delta$ \\
\midrule
\rowcolor{oursrow} Ours (consolidation) & 91.3 & -- \\
\midrule
\multicolumn{3}{l}{\textit{Initialization}} \\
\quad Start from the student & 71.4 & $-$19.9 \\
\quad RS-SFT initialization & 75.1 & $-$16.2 \\
\midrule
\multicolumn{3}{l}{\textit{Training queries}} \\
\quad Random queries & 77.4 & $-$13.9 \\
\quad Random queries, 10$\times$ pool & 83.5 & $-$7.8 \\
\quad Without error replay & 85.3 & $-$6.0 \\
\midrule
\multicolumn{3}{l}{\textit{Loss aggregation}} \\
\quad Global token mean & 60.7 & $-$30.6 \\
\quad Token mean per route & 76.0 & $-$15.3 \\
\quad Token-share sampling & 88.3 & $-$3.0 \\
\quad Dynamic sampling & 89.2 & $-$2.1 \\
\quad Spread normalization & 78.8 & $-$12.5 \\
\midrule
\multicolumn{3}{l}{\textit{Objective}} \\
\quad Without tail bucket & 74.3 & $-$17.0 \\
\quad Overlap tokens only & 75.5 & $-$15.8 \\
\quad Sampled token only & 72.6 & $-$18.7 \\
\midrule
\multicolumn{3}{l}{\textit{Preservation}} \\
\quad Without self teacher & 60.8 & $-$30.5 \\
\bottomrule
\end{tabular}
\caption{Ablations of the consolidation stage. Each row changes one component; $\Delta$ is the change in Rec. relative to ours.}
\label{tab:ablation}
\end{table}

\subsection{Diagnosis}
\label{sec:diagnosis}

\paragraph{Merging alone.}
Weight averaging recovers 74\% of the specialist's gain on business understanding and about a third on instruction following, perception, and domain reasoning, but falls below the student on general reasoning and agentic business tasks (Table~\ref{tab:main}).
Interpolating between the student and each specialist explains this pattern (Figure~\ref{fig:diag}a): business understanding gains most of its improvement close to the student, whereas general reasoning and agentic business tasks gain little, or even lose, until the weights approach their specialists.
Merging thus places the student close to some specialists but not others, so it serves as a starting point rather than a consolidation method on its own.

\paragraph{Refinement.}
After consolidation, general reasoning is the only capability below $\rho$ on the validation set, and from merging onward it remains farthest from its specialist in both overlap and score (Figure~\ref{fig:diag}b).
Continuing distillation from $\theta_1$ barely moves it (Table~\ref{tab:refine}), and its overlap with the general specialist stays near 77\%.
Merging the specialist into $\theta_1$ raises this overlap to 86\% and lifts general reasoning to the specialist level, but lowers the overlap with the domain specialist to 78\%, below $\tau$, while the other specialists stay above 82\%.
Without bringing the domain specialist back, domain reasoning falls to 96.6\%, and bringing back a specialist above $\tau$ instead does not prevent this drop.
Letting all specialists teach again limits this drop but lowers recovery to 88.0\%, as it no longer uses $\theta_1$ to preserve the capabilities that consolidation has already recovered.
Only our refinement keeps every capability at 98.9\% of its specialist or above.

\paragraph{Error replay.}
When replay starts, the overlap on general reasoning drops briefly as harder queries enter the batch and then recovers (Figure~\ref{fig:diag}c).
Its error pool shrinks from about 160 to 105 as new queries enter, so replayed errors are corrected rather than accumulated.
Yet this route retains the largest pool, signaling its lag (Appendix~\ref{app:diag}).

\begin{table}[t]
\centering
\footnotesize
\setlength{\tabcolsep}{3.5pt}
\begin{tabular}{lrrrr}
\toprule
Variant & Gen. & Dom. & Min. & Rec. \\
\midrule
No refinement ($\theta_1$) & 96.9 & 100.2 & 96.9 & 91.3 \\
\midrule
Without re-merging & 97.0 & 100.1 & 97.0 & 90.8 \\
Lagging specialist only, no self & 97.2 & 99.6 & 97.2 & 91.6 \\
Without bringing back & 100.0 & 96.6 & 96.6 & 85.0 \\
Bringing back a wrong specialist & 99.8 & 96.7 & 96.7 & 86.0 \\
All specialists teach again & 99.2 & 97.8 & 97.8 & 88.0 \\
Re-merge all, redo consolidation & 96.6 & 99.6 & 96.6 & 88.7 \\
\midrule
\rowcolor{oursrow} Ours (full recipe) & 99.6 & 98.9 & 98.9 & 97.0 \\
\bottomrule
\end{tabular}
\caption{Refinement variants, each run for 10 updates. Gen., Dom., and Min. are normalized scores on general reasoning, domain reasoning, and the weakest capability; Rec. is the recovery rate.}
\label{tab:refine}
\end{table}



\section{Conclusion}
\label{sec:conclusion}
We study how to consolidate specialists that have been post-trained far from a shared student back into a single model.
Our recipe initializes the student by merging, distills it on queries that the specialists solve but the student fails while replaying its errors, weights every specialist equally in the loss, and anchors training with a frozen self teacher, followed by a targeted refinement of capabilities that still lag.
With six multimodal specialists, the student recovers 97.0\% of their average gain, compared with 79.8\% for the strongest prior method, and closely approaches every specialist while surpassing one of them.
In the open five-domain setting, it likewise reaches 97.5\% recovery versus 94.1\% for IM-MOPD.
We also share practical observations on reading training signals, allocating the training budget, and configuring refinement, which we hope serve as a guide for practitioners who consolidate specialists at scale.

\newpage
\section*{Limitations}
Our study assumes that all specialists branch from the same student and share its architecture, which the merged initialization relies on, so consolidating heterogeneous teachers is left to future work.
Since the specialists are the same size as the student, we also do not study distilling larger specialists into a smaller student.


\bibliography{custom}

\newpage
\appendix

\section{Method Details}
\label{app:method}

Algorithm~\ref{alg:recipe} summarizes both stages of the recipe, using the notation of Section~\ref{sec:method}.

\begin{algorithm}[H]
\caption{Branch-and-consolidate recipe}
\label{alg:recipe}
\begin{algorithmic}[1]
\Require student $\theta_s$; specialists $\{\pi_k\}_{k=1}^{K}$ with parameters $\{\phi_k\}_{k=1}^{K}$; query distributions $\{\mathcal{D}_k\}_{k=1}^{K}$; thresholds $\rho$, $\tau$; steps $N_1$, $N_2$
\Statex \textit{Stage 1: consolidation}
\State $\theta_0 \gets \frac{1}{K+1}\bigl(\theta_s + \sum_{k} \phi_k\bigr)$ \Comment{Eq.~\eqref{eq:merge}}
\State $\pi_0 \gets$ frozen copy of $\pi_{\theta_0}$ \Comment{self teacher}
\For{$k = 1, \dots, K$}
  \State $\mathcal{Q}_k^{\mathrm{spec}} \gets$ queries where $\pi_k$ beats $\theta_s$
  \State $\mathcal{Q}_k^{\mathrm{self}} \gets$ random queries that $\theta_s$ solves
  \State $\mathcal{P}_k \gets \emptyset$ \Comment{replay pool}
\EndFor
\State $\mathcal{Q}^{\mathrm{self}} \gets \bigcup_{k} \mathcal{Q}_k^{\mathrm{self}}$
\State $\theta \gets \theta_0$
\For{$n = 1, \dots, N_1$}
  \State draw new queries from each $\mathcal{Q}_k^{\mathrm{spec}}$ and from $\mathcal{Q}^{\mathrm{self}}$
  \If{$n > N_1/4$}
    \State add replayed queries from each $\mathcal{P}_k$
  \EndIf
  \State roll out $\pi_\theta$ and route each $x$ to $k(x)$
  \State compute $\hat{\ell}_{x,t}$ \Comment{Eq.~\eqref{eq:topk}}
  \State update $\theta$ once on $\mathcal{L}$ \Comment{Eq.~\eqref{eq:agg}}
  \State add failed specialist-route queries to $\mathcal{P}_{k(x)}$ and remove solved ones
\EndFor
\State $\theta_1 \gets$ checkpoint selected on the validation set
\Statex \textit{Stage 2: refinement}
\State $\mathcal{G} \gets \{k : g_k(\theta_1) < \rho\}$ \Comment{Eq.~\eqref{eq:recovery}}
\State $\theta_1' \gets \frac{1}{1+|\mathcal{G}|}\bigl(\theta_1 + \sum_{k \in \mathcal{G}} \phi_k\bigr)$ \Comment{Eq.~\eqref{eq:remerge}}
\State $\mathcal{T} \gets \mathcal{G} \cup \{k \notin \mathcal{G} : \bar o_k(\theta_1') < \tau\}$
\For{$k = 1, \dots, K$}
  \If{$k \in \mathcal{T}$}
    \State route $k$: teacher $\pi_k$ on $\mathcal{Q}_k^{\mathrm{spec}}$
  \Else
    \State route $k$: frozen $\pi_{\theta_1}$ on $\mathcal{Q}_k^{\mathrm{self}}$
  \EndIf
\EndFor
\State $\theta \gets \theta_1'$
\For{$n = 1, \dots, N_2$}
  \State roll out, route, and update $\theta$ on $\mathcal{L}$ without replay \Comment{Eq.~\eqref{eq:agg}}
\EndFor
\State \Return checkpoint selected on the validation set
\end{algorithmic}
\end{algorithm}

\section{Experimental Details}
\label{app:setup}

\paragraph{Training.}
We train all parameters on 32 H100 GPUs without warmup, with one rollout per query and at most 16K generated tokens per model call.
After the first quarter of consolidation, each specialist route draws a fixed number of new queries per batch and up to the same number of replayed queries, depending on the size of its error pool.
The judge scores responses for query selection, error replay, and RS-SFT.
In Eq.~\eqref{eq:topk}, student probabilities are computed over the full vocabulary, teacher probabilities are floored at $10^{-8}$ and renormalized over the $b+1$ buckets, and the per-position loss is clipped at 10.

\paragraph{Validation and overlap.}
The validation set has 100 queries per capability, is frozen before training, and does not overlap the evaluation benchmarks.
Every trained method, including ablations, saves ten checkpoints at even intervals and is reported at the one selected on this set.
Overlap uses $b=32$ on a fixed probe batch of 256 queries covering all routes.

\subsection{Baselines}
\label{app:baselines}
All trained baselines sample queries at random from the same unfiltered candidate pools and run for as many updates as our consolidation and refinement combined, with our batch size and learning rate.
They start from the student, except IM-MOPD, which starts from $\theta_0$.

\paragraph{Weight merging.}
Uniform averaging equals $\theta_0$.
Task Arithmetic adds the sum of task vectors $\phi_k-\theta_s$ to the student with coefficient 0.5, and TIES additionally keeps the 20\% largest-magnitude entries of each task vector and resolves sign conflicts.

\paragraph{Offline distillation.}
RS-SFT samples two responses per query from the specialist, keeps those judged correct, using task-specific rubric thresholds for open-ended tasks, and fine-tunes the student on them.
Offline KD applies Eq.~\eqref{eq:topk} to one unfiltered specialist response per query.

\paragraph{On-policy distillation.}
Each variant adds its core mechanism to vanilla MOPD: Open-MOPD samples queries so that routes contribute equal shares of tokens, D$^3$-MOPD reweights domains dynamically with temperature 0.5, SF-MOPD projects teacher updates against an EMA slow model, and IM-MOPD adapts teacher contributions iteratively from a merged initialization.
Spread normalization in Table~\ref{tab:ablation} originally scales per-domain advantages; we apply the same clipped weight to the loss of each route, since our objective has no advantage.

\section{Fully Open Setting}
\label{app:open}

\paragraph{Setting.}
We follow the five-domain setting of IM-MOPD, in which all models, training data, and benchmarks are public.
The student is Qwen3-4B-Base fine-tuned on OpenThoughts3, and five specialists are branched from it: three for Medical (MedQA), Law (CaseHOLD), and Tool Use ($\tau^2$-Telecom) through supervised fine-tuning, and two for Finance (FinQA) and IF (IFBench) through reinforcement learning with verifiable rewards.
We choose this setting deliberately because three of its five specialists start below 70\% overlap with the student, between 66.2\% on Medical and 69.2\% on Tool Use (Table~\ref{tab:open_overlap}), which is the regime our recipe targets.
All methods use 100 updates with a batch of 128 prompts, one completion per prompt, a learning rate of $10^{-6}$, and at most 16{,}384 generated tokens.
Following IM-MOPD, we report pass@1 for Tool Use and avg@3 elsewhere, and compute the recovery rate of Eq.~\eqref{eq:rec} from the mean score over the five benchmarks.

\paragraph{Adapting the recipe.}
We keep every component and threshold of the recipe and run both stages within the same total budget as IM-MOPD, reserving the last 10\% of it for refinement.
Queries for Medical, Law, Finance, and IF are selected as in Section~\ref{sec:consolidation}, since their answers can be verified by rules.
Tool Use prompts are dialogue contexts without a verifiable answer for the next turn, so this route samples them at random, has no error replay, and draws its self queries at random as well.
Validation sets follow IM-MOPD.
Refinement targets Medical and Law, the two domains with $g_k < \rho$ on the validation set.
Re-merging their specialists raises the overlap with each of the other three specialists, all of which end above $\tau$ (Table~\ref{tab:open_overlap}), so these routes are taught by the frozen $\theta_1$.

\paragraph{Results.}
Table~\ref{tab:open_results} reports the results, with all baselines run under the same protocol.
Consolidation alone recovers 95.7\% of the specialists' average gain, compared with 94.1\% for IM-MOPD.
Re-merging raises the overlap with the medical and legal specialists from 73.9\% to 79.8\% and from 81.5\% to 85.4\% (Table~\ref{tab:open_overlap}), and refinement then improves MedQA from 74.7 to 77.3 and CaseHOLD from 71.0 to 71.6, while the other benchmarks change by at most 0.6 points.
The full recipe recovers 97.5\% and outscores IM-MOPD on all five benchmarks.

\begin{table}[t]
\centering
\small
\setlength{\tabcolsep}{3.5pt}
\begin{tabular}{@{}lrrrr@{}}
\toprule
Route & $\bar{o}_k(\theta_s)$ & $\bar{o}_k(\theta_0)$
      & $\bar{o}_k(\theta_1)$ & $\bar{o}_k(\theta_1')$ \\
\midrule
MedQA             & 66.24 & 71.26 & 73.86 & 79.83 \\
CaseHOLD          & 68.73 & 76.19 & 81.53 & 85.40 \\
FinQA             & 84.07 & 83.27 & 90.15 & 91.14    \\
IFBench           & 76.85 & 79.82 & 88.89 & 90.23   \\
$\tau^2$-Telecom  & 69.15 & 70.44 & 79.95 & 84.98   \\
\bottomrule
\end{tabular}
\caption{Top-32 overlap (\%) of each specialist with the initial
student $\theta_s$, the merged initialization $\theta_0$, the
consolidated student $\theta_1$, and the student re-merged with
the medical and legal specialists $\theta_1'$. Each route uses 100 fixed probe prompts. For $\tau^2$-Telecom, we use the first
agent turn. }
\label{tab:open_overlap}
\end{table}



\begin{table*}[t]
\centering
\small
\begin{tabular}{l*{7}{c}}
\toprule
\multirow{2}{*}{Method}
  & \multicolumn{5}{c}{Capability}
  & \multicolumn{2}{c}{Overall} \\
\cmidrule(lr){2-6}\cmidrule(lr){7-8}
  & MedQA & CaseHOLD & FinQA & IFBench & $\tau^2$ & Avg. & Rec. (\%) \\
\midrule
Student
  & 69.5 & 61.8 & 59.1 & 24.9 & 5.3 & 44.1 & 0.0 \\
Specialists
  & 80.7 & 73.4 & 75.0 & 59.4 & 66.7 & 71.0 & 100.0 \\
\midrule
Weight averaging
  & 71.7 & 66.5 & 69.0 & 38.0 & 10.5 & 51.1 & 26.1 \\
Offline KD
  & 64.5 & 61.6 & 72.8 & 54.6 & 36.8 & 58.1 & 51.8 \\
Vanilla MOPD
  & 67.2 & 63.9 & 73.9 & 58.6 & 52.6 & 63.2 & 71.0 \\
D$^{3}$-MOPD
  & 67.2 & 64.8 & \underline{74.9} & 57.6 & 56.1 & 64.1 & 74.3\\
SF-MOPD
  & 69.8 & 62.4 & 66.4 & 28.6 & 44.7 & 54.4 & 38.1\\
IM-MOPD
  & \underline{76.4} & 70.9 & 74.0 & 57.6
  & \underline{68.4} & 69.5 & 94.1 \\
\midrule
\multicolumn{8}{l}{\textit{Ours}} \\
\rowcolor{oursrow} Consolidation & 74.7 & \underline{71.0} & \textbf{75.0} & \textbf{59.4} & \textbf{69.3} & \underline{69.9} & \underline{95.7}\\
\rowcolor{oursrow} Full recipe & \textbf{77.3} & \textbf{71.6} & \underline{74.9} & \underline{58.8} & \textbf{69.3} & \textbf{70.4} & \textbf{97.5}\\
\bottomrule
\end{tabular}
\caption{Results in the open setting. Avg.\ is the mean score over the five benchmarks, and Rec.\ is the recovery rate in Eq.~\eqref{eq:rec} computed from it. Best results are in bold and second best underlined, with ties sharing the mark.}
\label{tab:open_results}
\end{table*}

\section{Additional Results}
\label{app:results}

\subsection{Per-Benchmark Results}
\label{app:per_bench}
Table~\ref{tab:per_bench} breaks general reasoning into its four public benchmarks, with MMMU-Pro evaluated in two settings.
After consolidation, most of the remaining gap comes from IMO-AnswerBench, where the consolidated model scores 73.25 against 85.00 for the specialist, and refinement raises it to 82.60.

\begin{table*}[t]
\centering
\small
\begin{tabular}{lcccccc}
\toprule
\multirow{2}{*}{Method} & \multirow{2}{*}{IMO-AnswerBench} & \multirow{2}{*}{MathVision} & \multicolumn{2}{c}{MMMU-Pro} & \multirow{2}{*}{LogicVista} & \multirow{2}{*}{Avg.} \\
\cmidrule(lr){4-5}
 & & & 10 options & Vision & & \\
\midrule
Qwen3.6-35B-A3B & 78.99 & 81.81 & 74.84 & 74.34 & 75.39 & 77.07 \\
Student & 69.70 & 85.03 & 73.18 & 71.45 & 76.96 & 75.26 \\
General specialist & 85.00 & 83.32 & 75.03 & 73.35 & 79.42 & 79.22 \\
\midrule
\multicolumn{7}{l}{\textit{Weight merging}} \\
Weight averaging & 70.20 & 84.05 & 72.85 & 70.65 & 74.25 & 74.40 \\
Task Arithmetic & 68.15 & 85.20 & 72.99 & 71.26 & 76.70 & 74.86 \\
TIES & 69.82 & 85.28 & 72.90 & 71.17 & 76.58 & 75.15 \\
\midrule
\multicolumn{7}{l}{\textit{Offline distillation}} \\
RS-SFT & 66.02 & 85.80 & 72.35 & 70.60 & 75.83 & 74.12 \\
Offline KD & 62.12 & 85.53 & 72.64 & 70.89 & 76.22 & 73.48 \\
\midrule
\multicolumn{7}{l}{\textit{On-policy distillation}} \\
Vanilla MOPD & 72.47 & 85.46 & 73.32 & 71.37 & 76.73 & 75.87 \\
Open-MOPD & 74.15 & 85.37 & 73.41 & 71.57 & 76.95 & 76.29 \\
D$^3$-MOPD & 74.82 & 85.46 & 73.32 & 71.47 & 77.23 & 76.46 \\
SF-MOPD & 67.55 & 84.86 & 73.37 & 71.64 & 77.18 & 74.92 \\
IM-MOPD & 78.77 & 84.09 & 74.20 & 72.50 & 78.29 & 77.57 \\
\midrule
\multicolumn{7}{l}{\textit{Ours}} \\
\rowcolor{oursrow} Consolidation & 73.25 & 84.47 & 74.05 & 71.50 & 80.54 & 76.76 \\
\rowcolor{oursrow} Full recipe & 82.60 & 84.90 & 74.40 & 73.10 & 79.50 & 78.90 \\
\bottomrule
\end{tabular}
\caption{Per-benchmark results on general reasoning. Avg. equals the general reasoning score in Table~\ref{tab:main}.}
\label{tab:per_bench}
\end{table*}

\subsection{Training Cost}
\label{app:cost}
Counted up to the selected checkpoint, consolidation takes 550 GPU hours and refinement 35 GPU hours on 32 H100 GPUs.

\section{Design Choice Details}
\label{app:design}
\begin{figure*}[t]
\centering
\includegraphics[width=\textwidth]{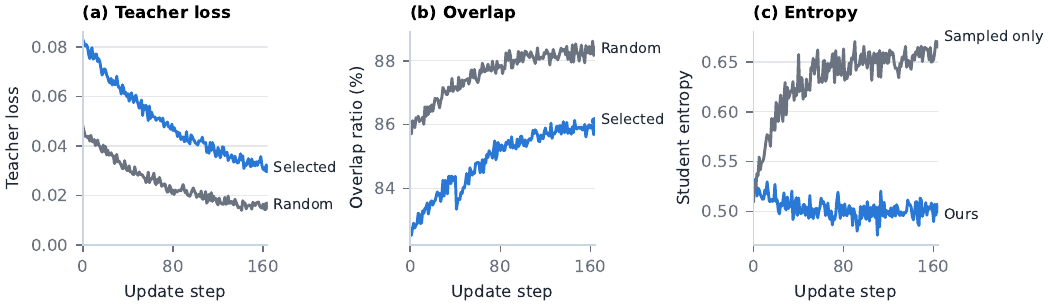}
\caption{Training curves behind three ablations in Section~\ref{sec:design}. (a, b) Random queries reach a lower teacher loss and a higher overlap than selected queries. (c) Using only the sampled token makes the student distribution more diffuse.}
\label{fig:design_appendix}
\end{figure*}

Table~\ref{tab:ablation_full} reports the per-capability scores of every ablation, including variants omitted from Table~\ref{tab:ablation}.
Merging with TIES instead of averaging loses 4.6 points, and mixing selected queries with queries that both models solve loses 5.7.
Forward KL loses 46.7 points, as it spreads the student's probability over more teacher candidates and raises its entropy.
Using 8 or 128 teacher candidates stays within 4 points of our 32, since the top 32 already cover most of the teacher's mass.
Doubling the self-teacher weight or using the self teacher only in the second half of training loses 9.0 and 10.5 points, whereas halving its weight changes little.
Halving or doubling the number of self queries per batch loses 1.2 and 4.3 points, much less than doubling the self-teacher weight, since the route average fixes its weight regardless of how many queries it draws.
Under the same budget, a halved learning rate or a larger batch with fewer updates loses 8.5 and 10.7 points; a larger batch gives more stable gradients, but halving the number of updates also halves the rounds of fresh rollouts, which costs more in on-policy distillation.
Training for two epochs doubles the updates and raises recovery to 94.0\%, since every pass draws fresh on-policy rollouts, yet refinement reaches 97.0\% at a small fraction of this extra cost (Appendix~\ref{app:cost}).
Figure~\ref{fig:design_appendix} shows the training curves behind three ablations in Section~\ref{sec:design}.

\begin{table*}[t]
\centering
\small
\begin{tabular}{lcccccccc}
\toprule
Variant & IF & Perc. & Agent & Bus. & Gen. & Dom. & Rec. (\%) & $\Delta$ \\
\midrule
\rowcolor{oursrow} Ours (consolidation) & 75.47 & 62.20 & 74.38 & 81.68 & 76.76 & 69.69 & 91.3 & -- \\
\midrule
\multicolumn{9}{l}{\textit{Initialization}} \\
\quad Start from the student & 74.90 & 61.35 & 73.95 & 81.20 & 76.10 & 68.40 & 71.4 & $-$19.9 \\
\quad RS-SFT initialization & 75.05 & 61.50 & 73.80 & 81.15 & 76.40 & 68.80 & 75.1 & $-$16.2 \\
\quad TIES merge as initialization & 75.98 & 62.49 & 72.84 & 82.23 & 75.33 & 70.32 & 86.7 & $-$4.6 \\
\midrule
\multicolumn{9}{l}{\textit{Training queries}} \\
\quad Random queries & 74.96 & 61.91 & 74.59 & 81.13 & 75.55 & 69.05 & 77.4 & $-$13.9 \\
\quad Random queries, 10$\times$ pool & 75.57 & 62.26 & 74.32 & 81.79 & 74.74 & 69.82 & 83.5 & $-$7.8 \\
\quad Without error replay & 75.26 & 62.08 & 74.30 & 81.46 & 76.36 & 69.43 & 85.3 & $-$6.0 \\
\quad Half selected, half both-correct queries & 75.64 & 62.30 & 74.38 & 81.87 & 74.86 & 69.91 & 85.6 & $-$5.7 \\
\midrule
\multicolumn{9}{l}{\textit{Loss aggregation}} \\
\quad Global token mean & 74.57 & 61.93 & 71.28 & 81.58 & 76.26 & 67.99 & 60.7 & $-$30.6 \\
\quad Token mean per route & 75.62 & 62.29 & 74.35 & 81.85 & 75.16 & 67.63 & 76.0 & $-$15.3 \\
\quad Token-share sampling & 75.80 & 62.35 & 73.95 & 81.95 & 75.70 & 69.78 & 88.3 & $-$3.0 \\
\quad Dynamic sampling & 75.28 & 62.09 & 74.20 & 81.47 & 76.86 & 69.82 & 89.2 & $-$2.1 \\
\quad Spread normalization & 75.64 & 62.30 & 74.27 & 81.87 & 75.44 & 67.98 & 78.8 & $-$12.5 \\
\midrule
\multicolumn{9}{l}{\textit{Objective}} \\
\quad Without tail bucket & 75.07 & 61.97 & 74.13 & 81.25 & 74.93 & 69.19 & 74.3 & $-$17.0 \\
\quad Overlap tokens only & 75.32 & 62.11 & 74.33 & 81.51 & 74.02 & 69.50 & 75.5 & $-$15.8 \\
\quad Sampled token only & 74.96 & 61.91 & 74.20 & 81.13 & 74.92 & 69.05 & 72.6 & $-$18.7 \\
\quad Forward KL & 74.00 & 61.35 & 74.44 & 80.08 & 72.46 & 67.84 & 44.6 & $-$46.7 \\
\quad Top-8 teacher candidates & 75.59 & 62.27 & 74.55 & 81.81 & 75.32 & 69.84 & 87.6 & $-$3.7 \\
\quad Top-128 teacher candidates & 75.45 & 62.19 & 74.39 & 81.66 & 76.39 & 69.67 & 89.3 & $-$2.0 \\
\midrule
\multicolumn{9}{l}{\textit{Preservation}} \\
\quad Without self teacher & 74.67 & 61.74 & 74.65 & 80.81 & 73.09 & 68.68 & 60.8 & $-$30.5 \\
\quad Self-teacher weight $\times 2$ & 75.29 & 62.10 & 74.55 & 81.48 & 75.37 & 69.46 & 82.3 & $-$9.0 \\
\quad Self-teacher weight $\times 0.5$ & 75.40 & 62.16 & 74.42 & 81.61 & 76.43 & 69.60 & 88.7 & $-$2.6 \\
\quad Self teacher in second half only & 75.35 & 62.13 & 74.45 & 81.55 & 74.91 & 69.54 & 80.8 & $-$10.5 \\
\quad Half self queries & 75.50 & 62.15 & 74.30 & 81.60 & 76.70 & 69.67 & 90.1 & $-$1.2 \\
\quad Double self queries & 75.40 & 62.18 & 74.35 & 81.55 & 76.25 & 69.52 & 87.0 & $-$4.3 \\
\midrule
\multicolumn{9}{l}{\textit{Optimization}} \\
\quad Half learning rate & 75.21 & 62.05 & 74.30 & 81.40 & 76.02 & 69.37 & 82.8 & $-$8.5 \\
\quad Batch size 512 & 75.29 & 62.10 & 74.32 & 81.48 & 75.24 & 69.46 & 80.6 & $-$10.7 \\
\quad Two epochs (2$\times$ updates) & 75.52 & 62.25 & 74.40 & 81.70 & 77.18 & 69.70 & 94.0 & $+$2.7 \\
\bottomrule
\end{tabular}
\caption{Per-capability scores of all ablations of the consolidation stage, including variants omitted from Table~\ref{tab:ablation}. All variants use the same training budget except two epochs.}
\label{tab:ablation_full}
\end{table*}

\begin{table*}[!t]
\centering
\small
\begin{tabular}{lccccccc}
\toprule
Variant & IF & Perc. & Agent & Bus. & Gen. & Dom. & Rec. (\%) \\
\midrule
No refinement ($\theta_1$) & 75.47 & 62.20 & 74.38 & 81.68 & 76.76 & 69.69 & 91.3 \\
\midrule
Without re-merging & 75.42 & 62.17 & 74.36 & 81.63 & 76.84 & 69.64 & 90.8 \\
Lagging specialist only, no self & 75.54 & 62.24 & 74.41 & 81.76 & 77.00 & 69.29 & 91.6 \\
Without bringing back & 75.04 & 61.95 & 74.20 & 81.22 & 79.22 & 67.20 & 85.0 \\
Bringing back a wrong specialist & 75.14 & 62.01 & 74.24 & 81.32 & 79.06 & 67.27 & 86.0 \\
All specialists teach again & 75.18 & 62.03 & 74.26 & 81.37 & 78.59 & 68.04 & 88.0 \\
Re-merge all, redo consolidation & 75.49 & 62.21 & 74.39 & 81.70 & 76.53 & 69.29 & 88.7 \\
\midrule
\rowcolor{oursrow} Ours (full recipe) & 75.60 & 62.90 & 74.30 & 80.90 & 78.90 & 68.80 & 97.0 \\
\bottomrule
\end{tabular}
\caption{Per-capability scores of the refinement variants in Table~\ref{tab:refine}, each run for 10 updates.}
\label{tab:refine_full}
\end{table*}

\section{Diagnosis Details}
\label{app:diag}

\begin{figure}[h]
\centering
\includegraphics[width=\columnwidth]{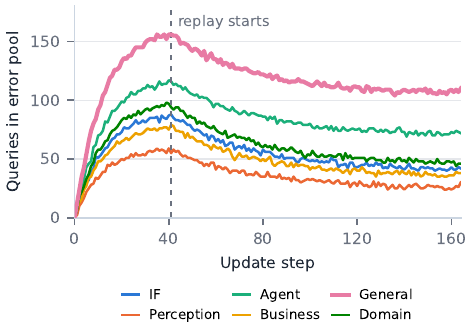}
\caption{Size of the error pool of each specialist route during consolidation.}
\label{fig:errorpool}
\end{figure}

\paragraph{Error replay.}
Figure~\ref{fig:errorpool} tracks each specialist route's error pool.
Each pool grows until replay starts, then shrinks even as new queries enter.
General reasoning retains the largest pool throughout, consistent with its lag after consolidation.

\paragraph{Refinement.}
Table~\ref{tab:remerge_overlap} lists the overlap that decides the teacher of each route.
Re-merging raises the overlap with the general specialist from 77.2\% to 86.0\% and lowers it for every other specialist, but only domain reasoning falls below $\tau$, so it is the only route besides general reasoning taught by its specialist.
Table~\ref{tab:refine_full} gives the per-capability scores of the refinement variants.

\begin{table}[h]
\centering
\footnotesize
\setlength{\tabcolsep}{3.5pt}
\begin{tabular}{lccl}
\toprule
Route & $\bar o_k(\theta_1)$ & $\bar o_k(\theta_1')$ & Teacher \\
\midrule
Instruction following & 86.1 & 83.5 & self \\
Perception & 90.2 & 88.4 & self \\
Agentic business & 84.7 & 82.6 & self \\
Business underst. & 86.9 & 84.2 & self \\
General reasoning & 77.2 & 86.0 & specialist (lagging) \\
Domain reasoning & 85.3 & 78.1 & specialist ($<\tau$) \\
\bottomrule
\end{tabular}
\caption{Overlap (\%) of each specialist with the consolidated student $\theta_1$ and the re-merged student $\theta_1'$, and the resulting teacher of each route in refinement ($\tau=80\%$).}
\label{tab:remerge_overlap}
\end{table}


\section{Practical Guide}
\label{app:monitor}
We collect practical observations from our runs on reading training signals and allocating the training budget during consolidation, and on configuring refinement.
We offer these heuristics as a guide for practitioners, though they have yet to be validated.
\paragraph{Reading overlap and loss together.}
Neither the teacher loss nor the overlap alone indicates how much a route still has to learn, so we read them jointly.
A low loss with a high overlap usually means that the student and the specialist already agree, as on random queries, so the route carries little training signal.
A high loss with a high overlap is the most useful case, since the two models share candidates but weight them differently, whereas a high loss with a low overlap means that the supervision is hard to exploit and the overlap should be raised first.
Merging shifts routes toward higher overlap, while query selection favors higher-loss examples. Together they place training in the most useful regime.

\paragraph{Merging is not a guaranteed gain.}
How much merging raises the overlap depends on how far each specialist has diverged from the student: in the industrial setting the gain ranges from 2.5 points on perception, whose specialist stays closest, to 6.5 points on general reasoning, whose specialist drifts farthest (Table~\ref{tab:routes}).
In the open setting, merging even lowers the overlap on FinQA, from 84.1\% to 83.3\% (Table~\ref{tab:open_overlap}).
When specialists remain close to the student, merging may add little, so we recommend measuring the initial overlap with and without merging before adopting it.

\paragraph{Diagnose with training signals, select with evaluation.}
Entropy, overlap, and loss help locate problems but do not rank checkpoints: random queries reach a lower loss and a higher overlap yet a lower recovery (Section~\ref{sec:design}).
We therefore select checkpoints only by task scores on the validation set.

\paragraph{Read every route separately.}
Global averages are dominated by routes with long responses, so a lagging route can hide behind a stable global loss.
We monitor the loss, the overlap, and the error pool of each route, and a pool that stops shrinking marks a capability that refinement will likely need to target.

\paragraph{Expect transient drops.}
When error replay starts, harder queries enter the batch and the overlap drops briefly before recovering (Figure~\ref{fig:diag}c), which reflects the change in data rather than unstable training.
In contrast, a KL to the self teacher that keeps rising instead of leveling off signals that the student is drifting away from capabilities it already has.

\paragraph{Training longer.}
When consolidation continues into a second epoch, the loss to the specialists keeps decreasing while the KL to the self teacher stays flat instead of rising, the entropy of the student stays on a plateau, and the overlap keeps increasing slowly.
Longer training thus keeps adding capability without drifting from the initialization, consistent with the gain of the two-epoch run in Appendix~\ref{app:design}.

\paragraph{Keeping the allocation fixed.}
Late in consolidation, some routes have saturated while others still improve.
An alternative to training longer is to reallocate a fixed budget across routes, lowering the loss weights or the number of queries of saturated routes and restoring them once these routes drop.
Such a scheme must judge saturation and drops from per-route signals such as the teacher loss, the overlap, and the student's entropy, which fluctuate from update to update and shift transiently when the data change, so a delayed or mistaken adjustment adds noise to training.
We instead keep the equal weights, do not adjust the query sampling of any route, and leave capabilities that still lag at the end to refinement.
We observe in our runs that a route that saturates early stops improving but does not drift while the other routes continue to learn, and that a route that degrades abruptly in the middle of training recovers over the following updates.
In both cases, the route stays under the full supervision of its specialist, which down-weighting it would weaken.
In our experience with six specialists, a longer schedule is affordable in industrial training, so trading budget for stability is safer than reallocating a fixed budget.
Internally, we have extended this recipe to dozens of specialists, where a schedule long enough for the slowest routes becomes costly and allocating the budget across routes grows in importance.
We leave this problem, together with which signals identify saturation and drops reliably, to future work.

\paragraph{When re-merging is unnecessary.}
Re-merging is needed when a lagging capability has a low overlap with its specialist, as general reasoning does in the industrial setting, where continuing distillation without re-merging barely moves the score (Table~\ref{tab:refine}).
In our other experiments, when the overlap is already high, on-policy distillation from $\theta_1$ with that specialist as the teacher is enough to recover the lagging capability, and the other capabilities forget little.
We thus check the overlap of each lagging capability before deciding whether to re-merge.

\paragraph{Setting the thresholds.}
We set $\rho=0.5$ and $\tau=0.8$ empirically and do not tune them.
The recovery threshold only needs to flag capabilities that lag far behind, since $g_k$ is computed on a small validation set and is used as a coarse signal.
The overlap threshold should separate specialists whose overlap drops sharply after re-merging from those that stay close to the student.

\end{document}